# Reliable Extraction of Clinical Follow-Up Instructions: A Hybrid Neural-Symbolic Pipeline

Michal Laufer[1], Yehudit Aperstein[2], Alexander Apartsin[3]

[1]Bar-Ilan University, Ramat Gan, Israel

[2]Afeka College of Engineering, Tel Aviv, Israel

[3]Holon Institute of Technology, Holon, Israel

**Abstract**

**Objective.** Outpatient notes carry follow-up instructions pairing actions with future times ("MRI brain in two weeks"). Extracting (action, date) pairs supports scheduling and audit, but generative extractors miss the date because linking and arithmetic are implicit in decoding. We test a hybrid neural-symbolic pipeline against direct generation.

**Methods.** We define TestSpecification and TimeSpecification entities and a ScheduledFor relation. BioBERT feeds BIO tagging and a biaffine linker; entities are canonicalized via a 28-action ontology and times normalized to day offsets deterministically. We evaluate on a 2,000-note synthetic outpatient corpus with action-disjoint splits (18 train, 6 OOV-test) against zero-shot GPT-4o-mini and LoRA-fine-tuned LLaMA-3 8B with note-level bootstrap 95% CIs.

**Results.** On 259-note seen and OOV splits the hybrid pipeline achieves Test-Time Pair F1 of 0.997 and 0.986 with 0.00-day MAE. Baselines reach high action F1 (LLaMA-3 0.992; GPT-4o-mini 0.963 seen) but Pair F1 stays at 0.51-0.57 (LLaMA-3) and 0.53 (GPT-4o-mini), CIs non-overlapping with the hybrid.

**Conclusion.** Separating learned entity extraction from deterministic date arithmetic outperforms generation on this benchmark, generalizes to held-out actions, and exposes failure modes. Transfer to real EHR notes is the next validation; a first-pass realism check is in Limitations.

**Keywords:** clinical natural language processing; information extraction; temporal expression normalization; relation extraction; BioBERT; follow-up instructions; synthetic clinical data

## 1. Introduction

Electronic health records contain a large volume of free-text documentation that records clinical plans, diagnostic uncertainty, rationale, and patient-specific instructions. Clinical information extraction has therefore become a central problem in biomedical informatics, with applications spanning cohort discovery, quality measurement, decision support, registry construction, and downstream validation of structured EHR fields [1]. Follow-up instructions are a particularly consequential subset of this information. A note may say that a patient should obtain an MRI in two weeks, repeat laboratory testing in three months, schedule physical therapy within five weeks, or return only if symptoms worsen. Failure to follow up on such instructions has been documented as a patient-safety issue in ambulatory care, with abnormal results and recommended actions sometimes going unaddressed across care transitions [23].

The technical challenge is not simply named entity recognition. A clinically useful extraction must recover a structured item, such as {"action": "MRI Brain", "period_text": "in 2 weeks", "period_date": "2026-01-24"}, where period_date is computed relative to the encounter date. This requires recognizing action spans, recognizing temporal expressions, linking the correct time to the correct action, and normalizing relative dates. The difficulty increases when notes contain multiple actions, shorthand expressions such as 3 mos or q6mo, historical temporal distractors, and plan sections whose structure varies across specialties and authors.

Prior clinical NLP has established strong foundations for concept extraction and temporal reasoning. Systems such as cTAKES operationalized modular clinical concept extraction for EHR text [2], while the i2b2 temporal challenge, THYME corpus, and Clinical TempEval tasks formalized event and time extraction in clinical narratives [3], [4], [5], [6]. Transformer encoders adapted to biomedical or clinical domains, including BioBERT, ClinicalBERT, PubMedBERT, and GatorTron, further improved task-specific extraction and relation modeling [7], [8], [9], [10], [11]. At the same time, large language models have made zero-shot or few-shot structured extraction from clinical notes feasible, but recent evaluations continue to report variability in strict format adherence, task-specific accuracy, and governance requirements [12], [13].

This work studies a narrower but operationally important question: for follow-up instruction extraction, is it better to ask a generative model to directly output action-date JSON, or to decompose the task into learned semantic extraction and deterministic date normalization? The repository evidence suggests that the decomposed design is promising on a controlled synthetic benchmark. The paper's central claim is therefore intentionally scoped: on synthetic, stress-varied outpatient notes, separating semantic span/link extraction from calendar arithmetic improves action-date reliability compared with the included direct generative baselines.

## Contributions

1. A structured information-extraction formulation for clinical follow-up instructions, defined by two entity types (TestSpecification, TimeSpecification) and an explicit ScheduledFor relation, with a strict set-level end-to-end correctness criterion over canonical (action, day-offset) pairs.
2. A hybrid neural-symbolic pipeline combining a shared BioBERT encoder, BIO entity tagging, biaffine relation extraction with a none option, ontology canonicalization, and deterministic temporal normalization, designed so that calendar arithmetic is never delegated to a learned model.
3. A 2,000-note synthetic outpatient corpus generated from controlled structured skeletons over five specialties, with explicit stress factors (multi-action notes, shorthand temporal forms, historical distractors, proximity traps, plan-section variation) and exact character-level ground truth.
4. A comparative evaluation against zero-shot GPT-4o-mini and LoRA-fine-tuned LLaMA-3 baselines on both an in-distribution (seen) test split and an action-disjoint out-of-vocabulary (OOV) test split, with note-level bootstrap 95% confidence intervals.
5. An error analysis showing that generative models often identify the correct TestSpecification label but fail on full-pair reconstruction because linking and temporal normalization are implicit in decoding.

## Statement of Significance

Problem or Issue

Outpatient clinical notes encode follow-up instructions ("MRI in 2 weeks") in free text alongside multiple actions, distractors, and shorthand. End-to-end LLM extraction can identify the action but commonly fails to reconstruct the complete scheduled instruction because linking and temporal arithmetic are implicit in generation and not exposed to inspection.

What is Already Known

Clinical NER systems (cTAKES, BioBERT, ClinicalBERT) handle medical concept extraction. Rule-based temporal-expression normalizers (HeidelTime, SUTime) handle relative-date arithmetic. LLM-based structured extraction is feasible but error-prone on relation and arithmetic components.

What this Paper Adds

A structured-information-extraction pipeline with explicit ScheduledFor relation linking and deterministic temporal normalization. On a controlled benchmark, the pipeline achieves end-to-end Test-Time Pair F1 of 0.997 (seen-test) and 0.986 (action-disjoint OOV-test) with 0.00-day mean date error, while LoRA-fine-tuned LLaMA-3 and zero-shot GPT-4o-mini reach 0.51 Pair F1 despite near-perfect action recognition.

Who would benefit from the new knowledge in this paper

Clinical informatics teams building EHR scheduling auditors, care-coordination tools, and structured-field validators; researchers studying neural-symbolic decomposition for safety-relevant clinical extraction.

## 2. Related Work

### 2.1 Clinical Information Extraction

Clinical information extraction aims to convert unstructured narrative text into structured representations usable for research and care operations. A broad review by Wang et al. describes clinical IE systems as pipelines that may include tokenization, syntactic processing, named entity recognition, concept identification, relation extraction, and context handling [1]. cTAKES is a prominent open-source clinical NLP system that extracts clinical concepts and attributes from free-text notes using a modular architecture [2]. The most direct prior work on the present task is Yetisgen-Yildiz et al., who built a text-processing pipeline to extract follow-up recommendations from radiology reports, demonstrating that scheduled-follow-up extraction can be operationalized as a sentence-level recommendation-detection problem followed by structured-field extraction [24]. Their work focuses on imaging reports rather than outpatient visit notes and on detecting whether a recommendation is present rather than recovering the linked time and normalized schedule; our task formulation extends this line to a structured entity-relation representation with deterministic temporal normalization.

The present task differs from standard concept extraction in two ways. First, the action to be scheduled is often a procedure, test, consult, or behavioral instruction rather than a diagnosis or medication mention. Second, extracting the action alone is insufficient: the system must link it to the temporal expression that determines when the action should occur. This places the task at the intersection of named entity recognition, relation extraction, and temporal normalization.

### 2.2 Temporal Information Extraction in Clinical Text

Clinical temporal extraction has been studied through shared tasks and corpora. The 2012 i2b2 challenge evaluated temporal relations in clinical text, including events, time expressions, and temporal links [3]. The THYME project developed clinical temporal annotation guidelines and corpora, emphasizing the specific demands of clinical narratives such as document sections, historical mentions, and event anchoring [4]. Clinical TempEval extended this line through SemEval tasks that evaluated systems on clinical temporal expression and relation extraction [5], [6].

These resources focus on general temporal structure in clinical narratives. Our task is narrower and more action-oriented: identify future follow-up actions and normalize their execution dates relative to a visit date.

This narrower target is clinically useful but also creates new evaluation requirements. A model may correctly detect a temporal phrase and still fail if it links that phrase to the wrong action or if it produces the wrong calendar date.

### 2.3 Temporal Normalization and Rule-Based Date Handling

Temporal expression recognition and normalization have a long tradition of rule-based and hybrid systems built around the TIMEX3 markup standard for time, duration, and set expressions. HeidelTime is a rule-based temporal tagger that extracts and normalizes temporal expressions across multiple domains and languages [14], and SUTime provides a library that recognizes and normalizes time expressions using compositional rules over part-of-speech and lexical features [15]. Clinical temporal extraction work emphasizes normalization as a foundational step for downstream temporal reasoning [16]. We implement deterministic normalization with the Python dateparser library, which provides a relative-time grammar in the same family.

The key design choice is to never ask the neural model to perform calendar arithmetic. The neural components identify spans and links; deterministic logic maps phrases such as "within five weeks" or "11 mos" to ISO dates against a fixed visit-date anchor. This follows a broader reliability-first pattern in clinical decision support: reserve learned models for ambiguous semantic interpretation, and use deterministic modules for arithmetic or schema-constrained transformations wherever possible. A practical advantage is that a wrong date has a single auditable cause, either a wrong linked time phrase or a wrong normalization, both of which are inspectable without re-running the model.

### 2.4 Domain-Specific Biomedical and Clinical Encoders

BERT established a general pretraining and fine-tuning framework for many NLP tasks [7]. Domain-specific pretraining then became important in biomedical and clinical NLP. BioBERT continues pretraining on PubMed abstracts and PMC full text and improves biomedical NER, relation extraction, and question answering [8]. ClinicalBERT adapts BERT to clinical notes and demonstrates the value of clinical-domain text for modeling EHR narratives [9]. PubMedBERT trains from scratch on biomedical literature and argues that in-domain vocabulary and pretraining matter for biomedical tasks [10]. GatorTron scales clinical language modeling using a large corpus of de-identified clinical text and evaluates across clinical NLP tasks including concept extraction and relation extraction [11].

Our use of BioBERT is therefore conservative rather than novel by itself. The novelty is not the encoder; it is the task-specific decomposition into action spans, time spans, action-time links, and deterministic date normalization for follow-up instructions.

### 2.5 Span-Based and Biaffine Relation Modeling

Relation extraction often benefits from explicit span-pair modeling. SpERT frames joint entity and relation extraction as a span-based transformer model, scoring candidate entities and relations over span representations [17]. Biaffine scoring, popularized in neural dependency parsing, provides a simple and effective way to score directed relations between two contextual representations [18]. The present architecture follows this family of ideas by representing action and time spans and scoring candidate action-time links, including a none option for actions without an explicit time.

### 2.6 LLM-Based Clinical Structured Extraction

Large language models have changed the baseline landscape for clinical information extraction. Agrawal et al. showed that large language models can act as zero- and few-shot clinical information extractors, introducing benchmark tasks based on clinical text [12]. Huang et al. assessed ChatGPT for structured extraction from clinical notes and reported promising feasibility, while also underscoring the need for careful evaluation against curated data [13]. Recent work in the scientific-text domain demonstrates that fine-tuned LLMs can extract structured records under controlled schemas [25]; the clinical setting introduces additional constraints on auditability and on the reliability of arithmetic substitutions such as date normalization. Recent clinical foundation-model reviews emphasize schema adherence, privacy, governance, and evaluation designs that reflect health-system value rather than benchmark convenience [19]. Prompt-engineering studies further show that LLM extraction performance is sensitive to task instructions, annotation guidance, and few-shot examples [20].
These studies motivate using direct generative extraction as a serious baseline. However, follow-up action-date extraction places pressure on exact dates and linking correctness. The results in this repository suggest that direct generation can recover actions well but may suffer more on time extraction, linking, and date arithmetic. This makes the task a useful setting for comparing general-purpose generative extraction with task-specialized decomposed extraction.

### 2.7 Synthetic Clinical Data

Clinical NLP research is constrained by privacy, governance, and access barriers. Synthetic clinical text can reduce PHI risk and support controlled stress testing, but it does not replace real-world validation. Kweon et al. train a publicly shareable clinical LLM on synthetic clinical notes derived from public case reports and evaluate it against real-note tasks, illustrating one route for privacy-aware clinical NLP research [21]. A 2025 scoping review of synthetic health record generation highlights both promise and inconsistent evaluation practices across medical text, time series, and longitudinal data [22]. The present corpus should therefore be positioned as a controlled evaluation scaffold, not as evidence that the model is ready for real clinical deployment.

## 3. Methods

### 3.1 Task definition

Each input example consists of a clinical note $x$ and a visit date $v$. The intermediate structured representation contains two entity types and one relation. **TestSpecification** entities are spans corresponding to scheduled clinical actions (e.g., X-Ray, Blood Test, CT Scan, Cardiology Consult). **TimeSpecification** entities are spans corresponding to the timing of an intended follow-up (e.g., "in 2 weeks", "tomorrow", "in 6 months"). A **ScheduledFor(TestSpecification, TimeSpecification)** relation connects each scheduled clinical action to the timing phrase that specifies when it should occur. TestSpecification entities are mapped to canonical ontology labels and TimeSpecification entities are mapped to integer day offsets relative to $v$. The final normalized output is the set:

$$Y = \{(a_i, k_i)\}_{i=1}^{n},$$

where $a_i$ is a canonical TestSpecification label drawn from a closed ontology and $k_i \in \mathbb{Z}$ is a normalized day offset. The model returns an empty set when no scheduled follow-up is present. The strict end-to-end correctness criterion is set-level: a prediction is correct only when its canonical label and its integer day offset both match the gold pair. The task decomposes into four subtasks: entity extraction (TestSpecification and TimeSpecification spans), ScheduledFor relation extraction, ontology canonicalization, and deterministic temporal normalization. Figure 1 shows the system pipeline.

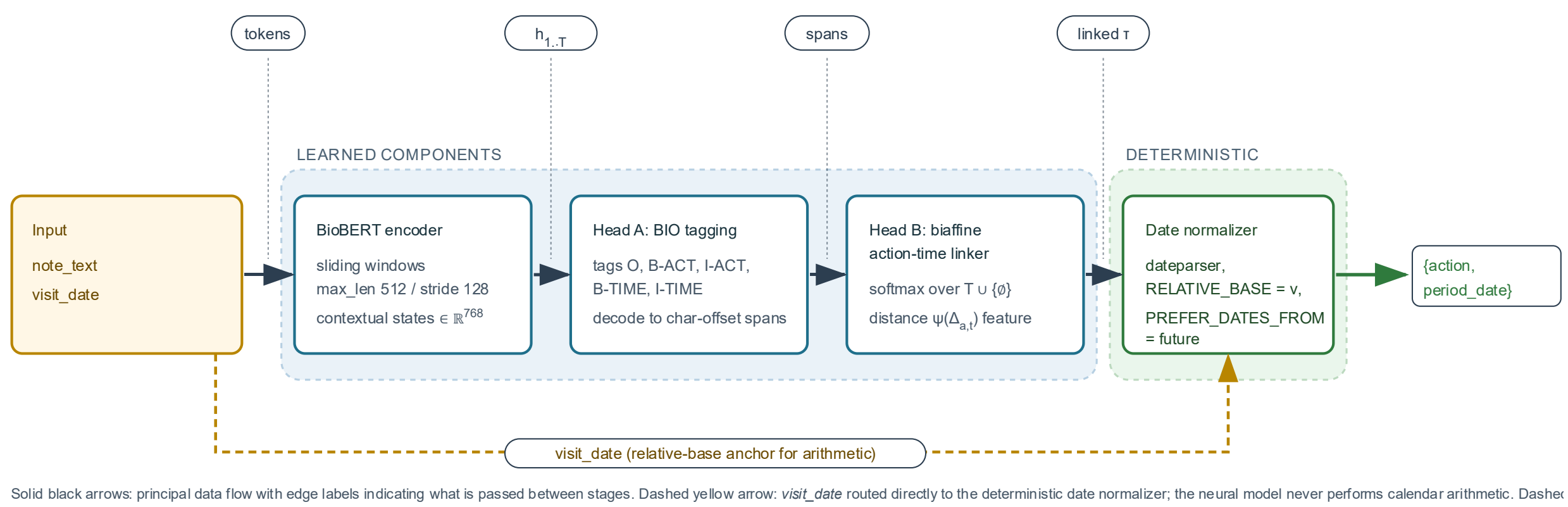

***Figure 1.*** *System overview. Solid arrows show the principal data flow. The note text is encoded by a shared BioBERT encoder over sliding windows of length 512 with stride 128; a BIO tagging head predicts TestSpecification and TimeSpecification entity spans from the contextual states; a biaffine relation head links each predicted TestSpecification span to a TimeSpecification span (or to a none option), producing a ScheduledFor relation; ontology canonicalization maps surface action mentions to one of the 28 canonical labels; a deterministic temporal normalizer maps the linked time phrase to an integer day offset relative to the visit date. The blue dashed enclosure groups the learned components (fine-tuned BioBERT plus the two heads); the green dashed enclosure marks the deterministic components. The dashed yellow arrow routes the visit_date directly to the normalizer as the relative-base anchor; the neural model never performs calendar arithmetic.*

### 3.2 Dataset and splits

We construct a 2,000-note synthetic outpatient corpus from controlled structured skeletons to avoid use of protected health information and to permit exact ground truth over entities, relations, and normalized timing. For each note, the generator first samples a clinical domain and scenario from one of five domains (orthopedics, cardiopulmonary medicine, gastroenterology, neurology, general medicine), a visit date in a two-year window (2024-01-21 to 2026-01-20), a target number of scheduled items (0, 1, or 2), TestSpecification labels from a closed set of 28 canonical action types, and TimeSpecification day offsets. Surface realization is varied separately from the structured skeleton: TimeSpecification expressions are rendered using multiple phrase families (numeric forms, written-number forms, abbreviations, fuzzy phrases, follow-up noun phrases), and notes vary in clinician voice, plan-header style, structural formatting, shorthand, abbreviations, and dictation-like artifacts. Non-target distractors are deliberately included (family history, prior procedures, symptom duration) to test discrimination between scheduled future instructions and historical context. Notes were generated with OpenAI gpt-4o-mini (default version as of November 2025); generation prompts, seed records, and the closed ontology are documented in the supplementary materials.

We use an **action-disjoint Seen/OOV split** to evaluate both in-distribution performance and generalization to action types not seen during training. The 28 canonical action types are partitioned by a numpy random shuffle (seed 123) into 18 training types, 4 OOV-validation types, and 6 OOV-test types. A note is assigned to a split if and only if all of its TestSpecification labels belong to that split's allowed action set. Train, OOV-validation, and OOV-test sets are therefore disjoint in both notes and in canonical action types. We additionally hold out a **seen-test** set drawn from notes whose action types are entirely within the training action set, to enable in-distribution comparison. After this partition the final splits contain 901 train notes, 100 seen-validation notes, 178 OOV-validation notes, 259 seen-test notes, and 259 OOV-test notes. We verified the leakage check at construction time: the train, seen-validation, and seen-test sets are disjoint by note, all 18 training action types are represented in training (minimum per-type count 22), and the OOV-test action types do not appear in any training note.

*Table 1. Benchmark and experimental splits. The corpus contains 2,000 synthetic outpatient notes with controlled item-level and span-level annotations. Ground truth is stored as canonical TestSpecification labels and integer day offsets, with exact character spans for both entity types. The action-disjoint Seen/OOV partition assigns notes to splits based on their TestSpecification types.*

| Property | Value |
|---|---|
| Total notes | 2,000 |
| Specialties (Orthopedic, Cardiopulmonary, Gastroenterology, Neurology, General Medicine) | 5 |
| Closed-set TestSpecification labels | 28 |
| Scheduled items per note (0 / 1 / 2) | 497 / 978 / 525 |
| Note length: min / median / max characters | 436 / 1,193 / 1,834 |
| Visit-date range | 2024-01-21 to 2026-01-20 |
| Training notes (18 seen action types) | 901 |
| Seen-validation notes | 100 |
| OOV-validation notes (4 held-out action types) | 178 |
| **Seen-test notes** (in-distribution) | **259** |
| **OOV-test notes** (6 held-out action types) | **259** |
| Split seed (numpy random shuffle) | 123 |

The 28 closed-set TestSpecification labels cover imaging (CT, MRI, MRI Brain, X-Ray, Echocardiogram, Cardiac MRI, Holter Monitor, Sleep Study, Abdominal Ultrasound), laboratory and procedural tests (Blood Test, Lipid Panel, Urinalysis, Stool Antigen Test, Breath Test, EMG, EEG, Stress Test, Pulmonary Function Test), procedures (Endoscopy, Colonoscopy, Joint Injection, Vaccination), specialty consults (Cardiology, Neurology, GI, Orthopedic), and rehabilitation/wellness referrals (Physical Therapy, Annual Physical). Stress factors include multi-action notes, historical temporal distractors, shorthand forms (3 mos, q6mo, x2w, RTC 3mo), proximity traps, and section ambiguity. The 2,028 gold action mentions are paired with 549 distinct temporal-expression surface forms; the most frequent expression accounts for only twelve mentions. All gold spans, canonical labels, and normalized day offsets are known by construction.

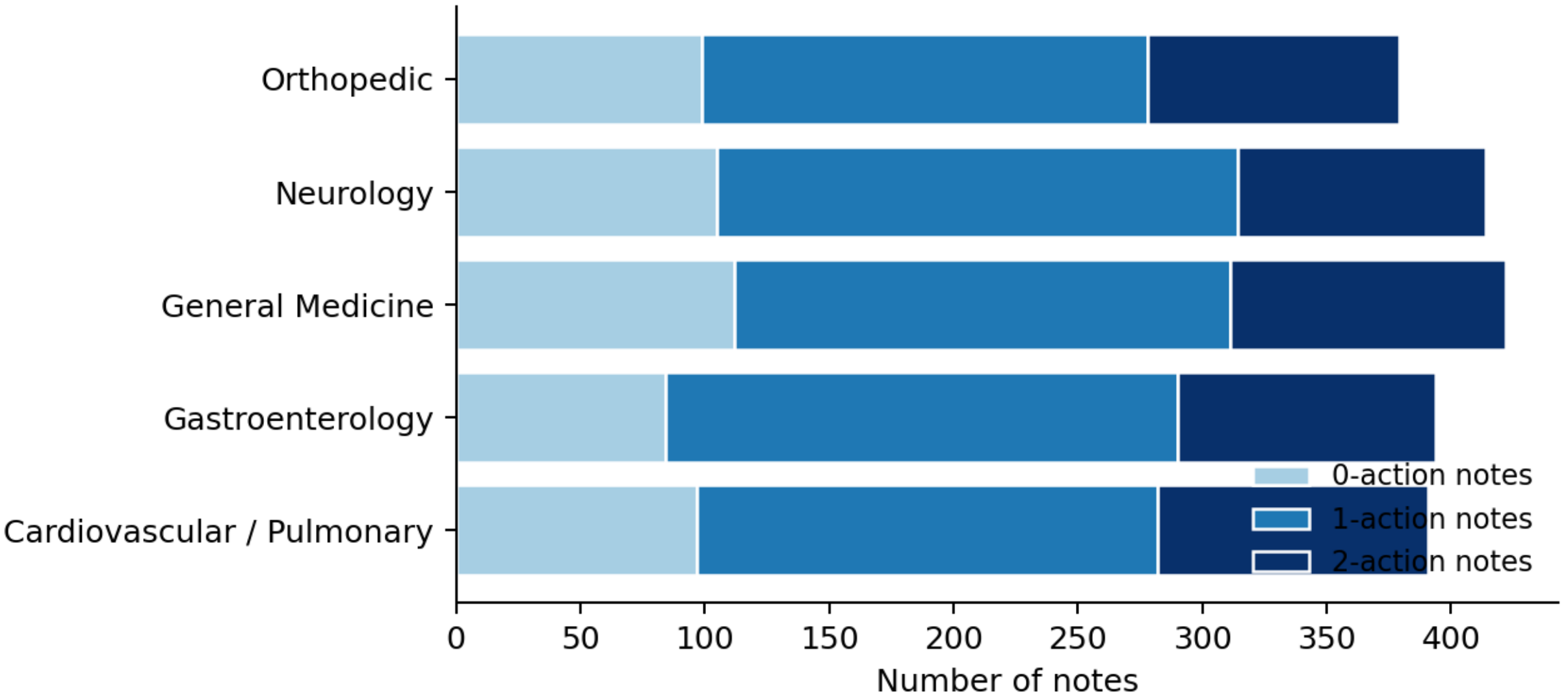

***Figure 2.*** *Dataset composition. Each bar shows one specialty stratum, segmented by the number of follow-up actions per note (0, 1, or 2). Specialty proportions are approximately balanced (379-422 notes per specialty), and the action-count distribution is consistent across specialties.*

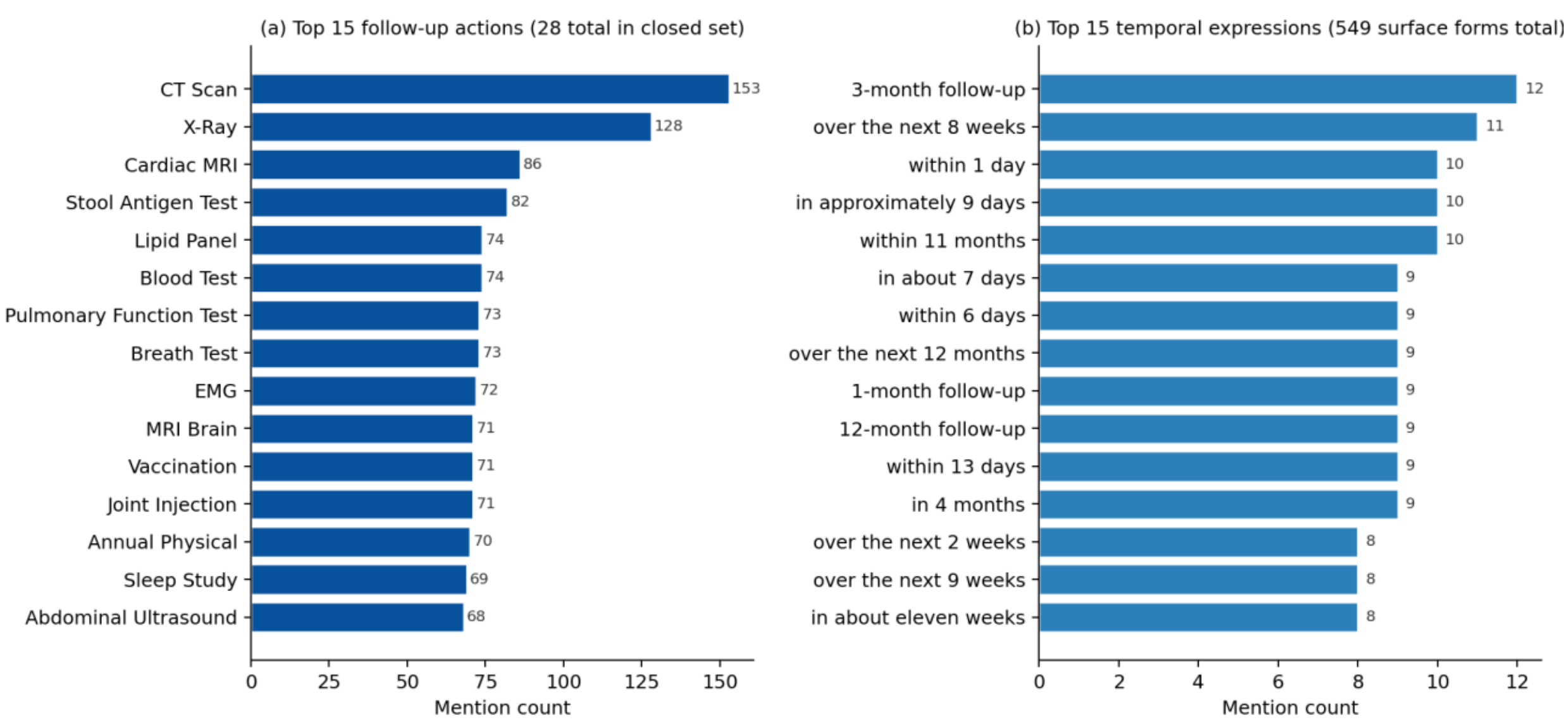

***Figure 3.*** *Vocabulary distributions in the corpus. (a) Top 15 follow-up action types out of 28 in the closed set; imaging and laboratory tests dominate, consults and rehabilitation appear in the long tail. (b) Top 15 temporal-expression surface forms out of 549 distinct forms among 2,028 gold mentions, illustrating the lexical diversity (the most frequent phrase has only twelve mentions).*

Stress factors are present at non-trivial proportions across the corpus: multi-action notes constitute 26.3%, zero-action notes 24.9%, historical-time mentions appear in roughly half of all notes, and the six plan-section variants are all represented (full per-factor counts are in the supplementary materials).

### 3.3 Hybrid neural-symbolic architecture

The proposed system shares a contextual encoder across two task heads (BIO entity tagging, biaffine ScheduledFor relation extraction) and delegates ontology canonicalization and temporal arithmetic to deterministic post-processors. We use the publicly released dmis-lab/biobert-base-cased-v1.1 checkpoint (768-dimensional hidden states) as the encoder, processing each note with sliding windows of maximum length 512 sub-word tokens and document stride 128; predictions from overlapping windows are merged at the character-span level. Let $h_1, \ldots, h_T \in \mathbb{R}^{768}$ denote the contextual token representations for a window.

*Entity extraction.* The first head predicts BIO tags over tokens using the label set $\{\mathrm{O}, \text{B-TEST}, \text{I-TEST}, \text{B-TIME}, \text{I-TIME}\}$, where TEST spans correspond to TestSpecification entities and TIME spans to TimeSpecification entities. The tagging loss is weighted cross-entropy with per-class weights downweighting the majority O class:

$$\mathcal{L}_{\mathrm{NER}} = -\sum_{t=1}^{T} w_{z_t} \log p(z_t \mid h_t).$$

At inference, contiguous BIO sequences are decoded into character-offset spans using the offset-mapping returned by the BioBERT tokenizer.

*Span representation.* For each detected entity span $s = (b, e)$, the model constructs a span representation by concatenating the start hidden state, the end hidden state, and a learned width embedding: $r_s = [h_b; h_e; \phi(e - b + 1)] \in \mathbb{R}^{1568}$. The width embedding $\phi$ maps span length to one of nine learned vectors.

*ScheduledFor relation extraction.* For each candidate (TestSpecification, TimeSpecification) pair, the relation extractor scores compatibility using a biaffine form augmented with a distance feature, $\mathrm{score}(a, t) = r_a^{\top} W r_t + u^{\top} [r_a; r_t; \psi(\Delta_{a,t})] + b$, where $\Delta_{a,t}$ is the absolute character-offset distance between the two spans, bucketized into learned 32-dimensional embeddings. For each TestSpecification span, the extractor applies a softmax over all candidate TimeSpecification spans plus a none option $\varnothing$ (a learned virtual span representing actions without an explicit time):

$$\mathcal{L}_{\mathrm{LINK}} = -\sum_{a \in A} \log p(t_a^{\star} \mid a, T \cup \{\varnothing\}).$$

The output ScheduledFor relation is written explicitly as ScheduledFor(test_entity_id, time_entity_id); this explicit relation is the key difference between the structured pipeline and direct generative baselines, because it exposes which entity was linked to which time expression before any normalization.

*Joint objective and training.* The total objective is $\mathcal{L} = \mathcal{L}_{\mathrm{NER}} + \alpha \mathcal{L}_{\mathrm{LINK}}$ with $\alpha = 1$. Training uses AdamW with batch size 16, learning rate $2\times10^{-5}$, and early stopping on validation loss; full hyperparameters are in the supplementary materials. Relation labels are derived from gold span pairs during training; at inference, the relation extractor operates on entity spans predicted by the tagging head.

*Ontology canonicalization and temporal normalization.* After relation extraction, TestSpecification surface forms (e.g., "CT test", "CT lab", "computed tomography") are mapped to canonical labels (CT Scan) through an editable ontology with an alias index; the ontology can be updated without retraining. Each linked TimeSpecification phrase is normalized to an integer day offset by a rule grammar over phrase families (tomorrow → 1, today → 0, in N weeks → $7N$, in N months → $30N$, in N years → $365N$; written-number and shorthand variants resolved before lookup). Unmatched phrases fall back to dateparser with RELATIVE_BASE = v and PREFER_DATES_FROM = "future". The neural model never performs calendar arithmetic; failures at the normalization stage are observable as parser-level errors and auditable against a fixed grammar.

### 3.4 Baselines

We compare against two generative baselines on identical inputs and the same held-out splits. **GPT-4o-mini zero-shot** receives the visit date, the note text, and the closed list of 28 allowed TestSpecification labels in a single system + user prompt, and is instructed to return a JSON array of {action, period_date} objects. Decoding uses temperature 0 and OpenAI structured-output mode (response_format = {"type": "json_object"}); no in-context examples are provided. For evaluation, the returned period_date is converted to a day offset relative to the visit date so that all systems are compared in the same normalized output space. **LLaMA-3 8B LoRA** uses meta-llama/Meta-Llama-3-8B as the base, loaded in 4-bit quantized form and adapted with LoRA adapters (rank 16, alpha 16, dropout 0) applied to query, key, value, output, and MLP projections. The base weights are frozen while LoRA adapters are trained on the same training split for 4 epochs at learning rate $2\times10^{-4}$, batch size 2 with gradient accumulation 8, and a maximum sequence length of 1024. At inference, the model is prompted with the same input format as the zero-shot baseline and decodes the JSON output, which is parsed and converted to day offsets for evaluation. Full prompts, ontology lists, and LoRA hyperparameters are documented in the supplementary materials.

### 3.5 Evaluation metrics

All systems are evaluated in a common normalized output space. Let the gold and predicted sets for a note be sets of canonical (TestSpecification, days_offset) pairs. We report three set-level F1 metrics and a calibration-style measure:

1. **TestSpecification F1.** Set-level F1 over canonical TestSpecification labels. A prediction is correct iff the predicted canonical label appears in the gold set.
2. **TimeSpecification Offset F1.** Set-level F1 over normalized integer day offsets. A prediction is correct iff the predicted offset appears in the gold set.
3. **Test-Time Pair F1 (end-to-end).** Set-level F1 over complete (TestSpecification, days_offset) pairs. A prediction is correct iff the entire pair matches a gold pair; this is the strict end-to-end metric. With $P$ the predicted set of pairs and $G$ the gold set, Precision $= |P \cap G|/|P|$, Recall $= |P \cap G|/|G|$, $F_1 = 2PR/(P + R)$.
4. **Time offset MAE on matched actions (days).** Among predictions whose canonical TestSpecification label matches a gold action, the mean of $|k_{\text{pred}} - k_{\text{gold}}|$ in days. The MAE is interpreted conditionally: if a model fails to extract the correct TestSpecification, that error is reflected in Pair F1 rather than in the MAE.

### 3.6 Statistical analysis

We report note-level bootstrap 95% confidence intervals for every metric in Table 2. For each metric we draw 1,000 bootstrap samples (random seed 123) from the held-out evaluation set, resampling notes with replacement and recomputing the metric on each sample; the 2.5th and 97.5th percentiles of the resulting distribution form the 95% confidence interval. All confidence-interval calculations are reproducible from the released per-system metric files using the analysis code distributed with the supplementary materials. We treat non-overlapping 95% confidence intervals between two systems as evidence of a significant difference at $p < 0.05$ on the corresponding metric.

## 4. Results

Table 2 reports point estimates and note-level bootstrap 95% confidence intervals for every metric on both the seen-test split (n=259 notes, action types in the training set) and the action-disjoint OOV-test split (n=259 notes, action types held out from training). Figure 4 visualizes the three F1 metrics; Figure 5 visualizes the time-offset MAE.

*Table 2. Seen and OOV evaluation results with note-level bootstrap 95% confidence intervals. Both test sets contain 259 notes. The seen-test set uses notes whose TestSpecification types are entirely within the 18 training action types; the OOV-test set uses notes whose TestSpecification types are entirely within the 6 held-out action types.*

| Model | Split | TestSpecification F1 | TimeSpecification Offset F1 | Test-Time Pair F1 | Time offset MAE (days) |
|---|---|---|---|---|---|
| BioBERT structured pipeline | Seen | 0.997 [0.990, 1.000] | 1.000 [1.000, 1.000] | **0.997 [0.990, 1.000]** | **0.00 [0.00, 0.00]** |
| BioBERT structured pipeline | OOV | 0.990 [0.979, 0.998] | 0.986 [0.972, 0.996] | **0.986 [0.972, 0.997]** | **0.00 [0.00, 0.00]** |
| LLaMA-3 8B LoRA | Seen | 0.992 [0.985, 0.998] | 0.511 [0.455, 0.563] | 0.510 [0.455, 0.562] | 21.31 [12.75, 31.00] |
| LLaMA-3 8B LoRA | OOV | 0.991 [0.980, 0.998] | 0.564 [0.509, 0.624] | 0.567 [0.512, 0.625] | 6.25 [2.56, 11.14] |
| GPT-4o-mini zero-shot | Seen | 0.963 [0.947, 0.977] | 0.530 [0.472, 0.582] | 0.528 [0.471, 0.581] | 23.65 [14.06, 34.12] |
| GPT-4o-mini zero-shot | OOV | 0.932 [0.909, 0.952] | 0.533 [0.474, 0.592] | 0.532 [0.474, 0.590] | 10.36 [4.38, 17.18] |

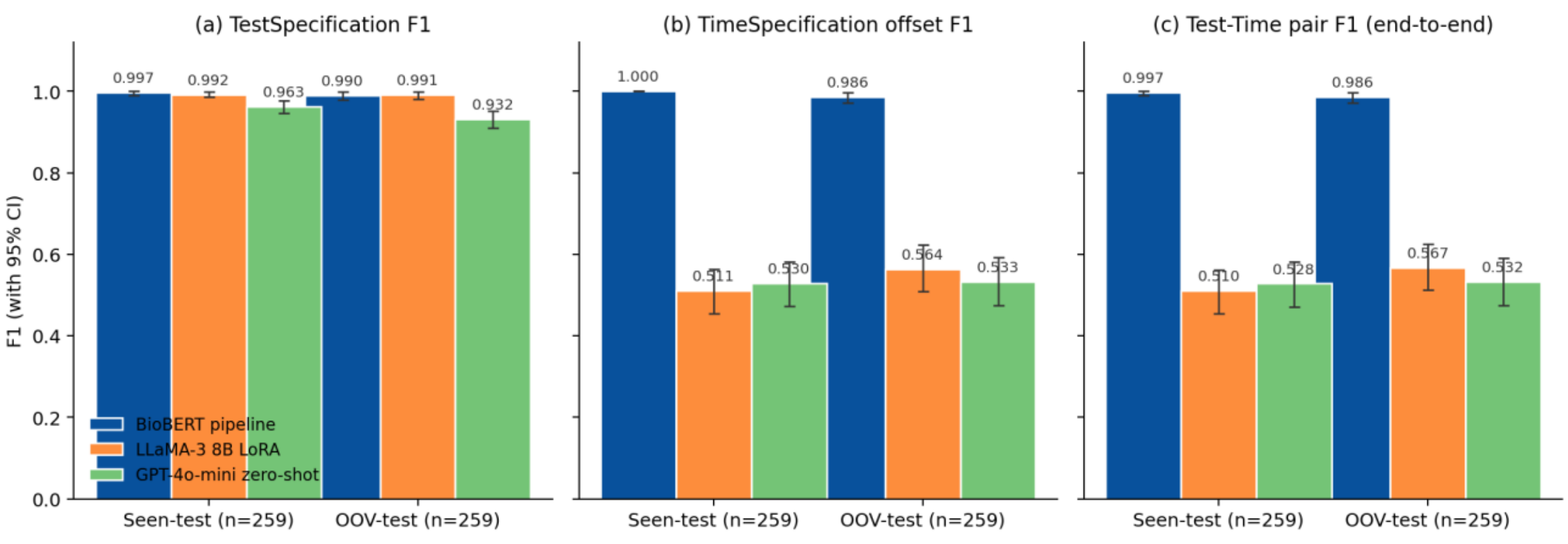

***Figure 4.*** *Held-out performance with note-level bootstrap 95% confidence intervals on the seen-test (in-distribution) and OOV-test (action-disjoint) splits. The structured pipeline reaches near-perfect F1 on all three metrics in both splits. Generative baselines achieve high TestSpecification F1 but their TimeSpecification Offset and Test-Time Pair F1 drop sharply, with CIs that do not overlap the BioBERT intervals on either split.*

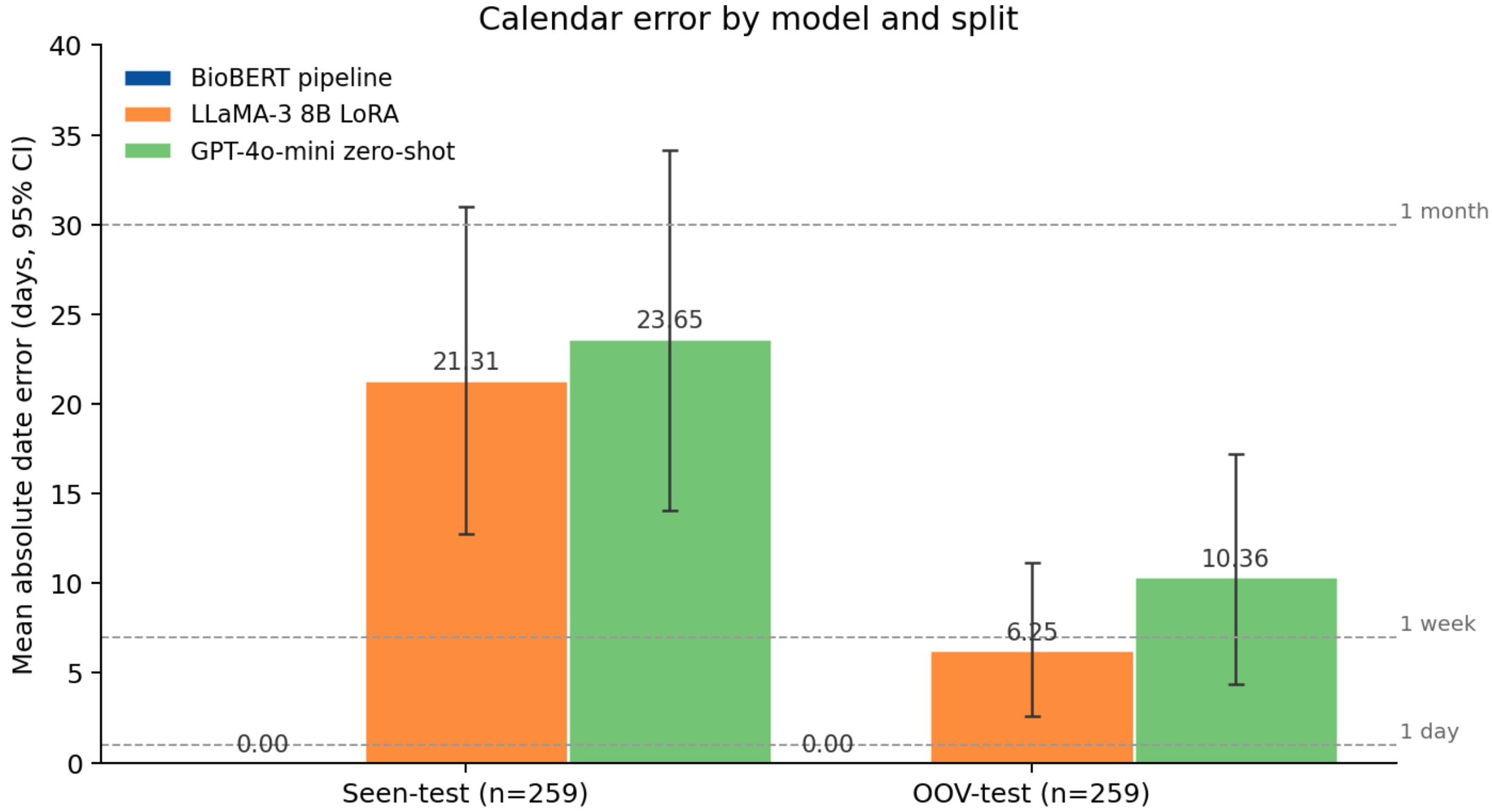

***Figure 5.*** *Mean absolute time-offset error in days on matched TestSpecification entities, with note-level bootstrap 95% confidence intervals. Reference lines mark one day, one week, and one month. The structured pipeline achieves 0.00-day MAE in both splits because every linked time phrase is normalized by the deterministic rule grammar. Generative baselines produce day-scale to month-scale arithmetic errors.*

The structured pipeline achieves near-perfect performance on both splits: Test-Time Pair F1 is 0.997 [0.990, 1.000] on the seen split and 0.986 [0.972, 0.997] on the OOV split. The modest reduction on OOV is expected because the action types themselves are excluded from training, yet the model still generalizes by detecting unseen TestSpecification surface forms via the BIO tagger and routing them through the same relation extractor and deterministic normalizer. The TimeSpecification Offset F1 is 1.000 on the seen split and 0.986 on OOV, indicating that the temporal-extraction component is essentially split-invariant. The pipeline's time-offset MAE is 0.00 days in both splits.

In contrast, both generative baselines achieve high TestSpecification F1 (LLaMA-3 0.992 / 0.991; GPT-4o-mini 0.963 / 0.932) but their Test-Time Pair F1 is approximately half: 0.510 / 0.567 for LLaMA-3 and 0.528 / 0.532 for GPT-4o-mini. The 95% confidence intervals for the structured pipeline's Pair F1 and Offset F1 do not overlap with the corresponding intervals of either generative baseline on either split, supporting a significant difference at $p < 0.05$ on these end-to-end metrics. The gap between near-perfect action recognition and ~0.51 Pair F1 reveals that the generative baselines correctly identify the clinical action label but fail to reconstruct the complete scheduled instruction.

### 4.1 Error analysis

The dominant error pattern across both generative baselines is a gap between TestSpecification identification and Test-Time pair reconstruction. This is most visible for LLaMA-3, where action recognition is near-perfect (F1 0.992 on seen) while pair reconstruction is ~0.51. Direct generation fails not because the model cannot name the action but because linking and temporal normalization are implicit in decoding and not exposed to inspection. Table 3 summarizes the qualitative error profile by model family.

*Table 3. Qualitative error profile by model family on the seen and OOV test splits.*

| Error category | BioBERT structured pipeline | LLaMA-3 8B LoRA | GPT-4o-mini zero-shot |
|---|---|---|---|
| Missed TestSpecification | Rare | Rare | Moderate |
| Hallucinated / unsupported TestSpecification | Very rare (ontology canonicalization) | Low (ontology-assisted prompt) | Possible in zero-shot generation |
| Wrong TimeSpecification surface form | Rare when linked span is correct | Common | Common |
| Wrong Test-Time relation | Rare; explicit relation extractor | Implicit and not inspectable | Implicit and not inspectable |
| Temporal arithmetic error | None; deterministic normalization | High | High |
| Auditability of failures | High: spans, links, normalized fields all inspectable | Low: direct JSON generation | Low: direct JSON generation |

Representative failures illustrate the temporal-arithmetic pattern: for "in approximately two months" (gold 60 days), GPT-4o-mini emitted a 304-day negative offset (year-boundary hallucination); for "ten mos" (gold 300 days), LLaMA-3 emitted 182 days (abbreviation misinterpretation); for "9-week follow-up" (gold 63 days), GPT-4o-mini emitted 60 days (small calendar mismatch). The structured pipeline routes these same phrases through the deterministic rule grammar, which maps them unambiguously to 60, 300, and 63 days. Additional worked examples are reported in the supplementary materials.

## 5. Discussion

**The decisive gap is on full-pair reconstruction, not action recognition.** Both generative baselines achieve high TestSpecification F1 (LLaMA-3 0.992 / 0.991; GPT-4o-mini 0.963 / 0.932) but their Test-Time Pair F1 is approximately half (0.510 / 0.567 and 0.528 / 0.532). Models that correctly name the clinical action fail to assemble the full schedule, because linking and temporal normalization are implicit in decoding and not exposed to inspection. The structured pipeline closes this gap by representing the ScheduledFor relation explicitly and delegating temporal arithmetic to a deterministic post-processor.

**Auditability of the structured pipeline.** The neural and symbolic components address qualitatively different failure modes. The neural model handles ambiguous semantic interpretation, deciding which span is a TestSpecification, which is a TimeSpecification, and which time belongs to which action; the deterministic normalizer handles a narrow well-specified arithmetic problem, mapping a phrase such as "11 mos" together with the visit date to an integer day offset. Combining the two creates an auditable boundary: any wrong end-to-end pair can be attributed either to a wrong entity span (inspectable as a predicted character span), or to a wrong ScheduledFor link (inspectable as a predicted relation tuple), or to a wrong normalization (inspectable by re-running the rule grammar on the predicted phrase). An end-to-end generative model that emits "2025-06-13" exposes none of these failure causes to inspection without reverse-engineering the decoder.

**Fine-tuning on in-distribution data does not close the gap.** The LLaMA-3 baseline is fine-tuned on the same training notes the structured pipeline uses and is given the closed ontology in its prompt, yet its Pair F1 remains at 0.510 on the seen-test split. The bottleneck is therefore not action-label recognition or ontology mismatch; it is the inability of unconstrained decoding to reliably reconstruct linked, normalized pairs. The TimeSpecification Offset F1 of 0.511 on seen-test and 0.564 on OOV-test for LLaMA-3 indicates the temporal-arithmetic component is the dominant failure mode. This is consistent with a generic argument: calendar arithmetic is a problem of computation, not of language, and is better delegated to a deterministic rule grammar than learned. The comparison reflects the specific LoRA configuration used here and would not necessarily hold for tool-using or constrained-decoding LLMs; nevertheless, it is informative that a fine-tuned 8B-parameter model with full access to the training distribution did not match a deterministic post-processor on the same time phrases.

**Generalization to held-out action types.** The structured pipeline retains near-perfect performance on the action-disjoint OOV-test split (Pair F1 0.986, MAE 0.00 days) despite never having seen the six OOV action types during training. The drop from seen to OOV is small (0.997 to 0.986 Pair F1) and is concentrated in TestSpecification F1 (0.997 to 0.990) rather than in Offset or arithmetic components. This indicates that the BIO tagger generalizes by surface-form features rather than by memorizing a closed action vocabulary, and that the ontology canonicalization stage can be updated to admit new action types without retraining the encoder. The generative baselines also generalize on action recognition (LLaMA-3 0.992 to 0.991, GPT-4o-mini 0.963 to 0.932) but their Pair F1 remains low on both splits, confirming that the OOV gap is not where the architecture difference matters.

**Scope of the LLM comparison.** The comparison with generative baselines is not a general indictment of LLMs for clinical extraction. LLMs may perform better with stronger prompts, tool use, constrained decoding, or retrieval; in particular, a tool-using LLM that delegates arithmetic to an external calculator would by construction recover most of the deterministic normalizer's advantage. The narrow conclusion is that, on this controlled benchmark, an architecture with explicit relation extraction and deterministic temporal normalization materially outperforms two reasonable end-to-end generative baselines on full-pair correctness and calendar arithmetic, with non-overlapping 95% CIs on the headline metrics in both in-distribution and OOV settings.

**Synthetic-data caveats.** Synthetic notes are valuable for controlled stress testing because the true action, time, and offset are known by construction; this enables exact-span metrics and the action-disjoint OOV evaluation that real-note evaluation cannot easily support. However, synthetic data can encode generator artifacts: consistent phrasing tics, a fixed closed ontology, predictable plan structure. The corpus partially mitigates this through ten style-feature axes, multiple plan-section variants, 549 distinct temporal-expression surface forms, and explicit stress factors. A model that overfits to the generator's surface lexicon would not be detectable from the present evaluation alone; transfer to real EHR documentation is the principal external-validity question.

### 5.1 Limitations and future work

The study has several limitations that bound the conclusions. *Synthetic evaluation:* the corpus is generated from controlled structured skeletons rather than drawn from real EHR documentation; generalization to real outpatient notes is the principal open question. *Limited action ontology:* the closed set covers imaging, laboratory and procedural tests, consults, and rehabilitation referrals, but does not include medication changes, conditional follow-up ("return if symptoms worsen"), patient self-care behavior, or instructions with ambiguous executor. *Single ontology and single normalization grammar:* we use one closed action ontology and one English-language temporal rule grammar; both are editable without retraining but neither has been validated outside the synthetic corpus. *No workflow validation:* the study does not assess clinician acceptance, EHR integration, alert behavior, or downstream scheduling outcomes.

*Realism check on real transcribed notes.* A first-pass annotation of the twenty highest-scoring real transcribed outpatient notes from the public MTSamples corpus (Apache-2.0; approximately 5,000 clinician transcription samples) found that 40% (19 of 48 identified follow-up items) map cleanly to the closed-set ontology, with the remaining 60% falling into categories the synthetic corpus does not represent: medication titration (25% of items), generic specialist follow-up appointments without a specified procedure (17%), closed-set actions recommended without a scheduled time (13%), specialist referrals outside the consult set (8%), and conditional or self-care instructions. This is direct empirical evidence for the ontology-coverage limitation above and motivates extending the entity schema and ontology before any real-EHR deployment. The 20-note annotation and per-item coverage analysis are released as supplementary materials.

Planned future work includes: (i) extending the TestSpecification ontology to medication management and conditional follow-up, with corresponding schema changes; (ii) encoder ablations across general BERT, ClinicalBERT, PubMedBERT, and GatorTron; (iii) per-stratum performance by specialty, note length, scheduled-item count, shorthand temporal forms, historical distractors, and proximity traps; (iv) a relation-extraction ablation against nearest-time, same-sentence, and action-order heuristics; (v) a temporal-normalization ablation against model-emitted day offsets with entity extraction held fixed; (vi) a clinician-annotated error taxonomy on a small de-identified real-note sample; and (vii) external evaluation on a de-identified institutional outpatient corpus or on a public clinical temporal-extraction benchmark adapted to ScheduledFor extraction.

## 6. Conclusion

We presented a hybrid neural-symbolic pipeline that formulates follow-up extraction as structured information extraction over TestSpecification and TimeSpecification entities linked by an explicit ScheduledFor relation, followed by ontology canonicalization and deterministic temporal normalization. On a 2,000-note synthetic outpatient benchmark with an action-disjoint Seen/OOV split design, the pipeline achieves Test-Time Pair F1 of 0.997 (seen-test) and 0.986 (OOV-test) with 0.00-day mean absolute time-offset error in both splits. Generative baselines (LoRA-fine-tuned LLaMA-3 8B and zero-shot GPT-4o-mini) achieve high TestSpecification F1 but only approximately 0.51 Pair F1, indicating that direct generation can name clinical actions but cannot reliably reconstruct the complete scheduled instruction. The architectural separation between learned semantic extraction and deterministic symbolic transformation exposes failure modes to inspection: any wrong pair can be attributed to a specific entity span, ScheduledFor link, or normalization step. The system is applicable to scheduling-auditor and care-coordination tools that operate downstream of structured EHR fields and require a trustworthy mapping from free-text plans to executable schedules. Transfer to real EHR notes, ontology extension to medication management and conditional follow-up, and a clinician-validated error taxonomy are the principal next steps before clinical deployment.

## Data and code availability

The synthetic dataset described in Section 3.2, the training and inference code, the per-system metric files (point estimates and reconstructed success counts), the analysis code that produces the confidence intervals, and the source files for every manuscript figure are publicly available under permissive licences at https://github.com/ApartsinProjects/MedFollow. A persistent DOI release on Zenodo will be deposited at acceptance and cited in the final version. Trained model checkpoints (the fine-tuned BioBERT weights and the LLaMA-3 LoRA adapter) are not redistributed in the public release because of repository-size constraints and are available from the corresponding author on reasonable request.